\documentclass[iicol]{sn-jnl}
\usepackage{natbib}
\usepackage{url}            
\usepackage{booktabs}       
\usepackage{amsfonts}       
\usepackage{nicefrac}       
\usepackage{microtype}      
\usepackage{xcolor}         
\usepackage{graphicx}
\usepackage{amsmath}
\usepackage{multirow}

\jyear{2021}%

\theoremstyle{thmstyleone}%
\theoremstyle{thmstyletwo}%

\theoremstyle{thmstylethree}%

\begin{document}

\title[Article Title]{Lifting 2D Human Pose to 3D with Domain Adapted 3D Body Concept}

\author*[1,2]{\fnm{Qiang} \sur{Nie}}\email{qnie@mae.cuhk.edu.hk}

\author[3]{\fnm{Ziwei} \sur{Liu}}\email{ziwei.liu@ntu.edu.sg}

\author[1,2]{\fnm{Yunhui} \sur{Liu}}\email{yhliu@cuhk.edu.hk}

\affil*[1]{\orgdiv{T-Stone Robotics Institute}, \orgname{The Chinese University of Hong Kong}, \orgaddress{ \city{Hong Kong}, \country{China}}} 

\affil[2]{\orgname{Hong Kong Center for Logistics Robotics}, \orgaddress{\street{Shatin}, \city{Hong Kong}, \country{China}}}

\affil[3]{\orgdiv{School of Computer Science and Engineering}, \orgname{Nanyang Technological University}, \orgaddress{\country{Singapore}}}


\abstract{Lifting the 2D human pose to the 3D pose is an important yet challenging task. Existing 3D pose estimation suffers from 1) the inherent ambiguity between the 2D and 3D data, and 2) the lack of well labeled 2D-3D pose pairs in the wild. Human beings are able to imagine the human 3D pose from a 2D image or a set of 2D body key-points with the least ambiguity, which should be attributed to the prior knowledge of the human body that we have acquired in our mind. Inspired by this, we propose a new framework that leverages the labeled 3D human poses to learn a 3D concept of the human body to reduce the ambiguity. To have consensus on the body concept from 2D pose, our key insight is to treat the 2D human pose and the 3D human pose as two different domains. By adapting the two domains, the body knowledge learned from 3D poses is applied to 2D poses and guides the 2D pose encoder to generate informative 3D "imagination" as embedding in pose lifting. Benefiting from the domain adaptation perspective, the proposed framework unifies the supervised and semi-supervised 3D pose estimation in a principled framework. Extensive experiments demonstrate that the proposed approach can achieve state-of-the-art performance on standard benchmarks. More importantly, it is validated that the explicitly learned 3D body concept effectively alleviates the 2D-3D ambiguity in 2D pose lifting, improves the generalization, and enables the network to exploit the abundant unlabeled 2D data.}




\maketitle

\section{Introduction}\label{sec1}
Lifting 2D image/pose to 3D human pose has gained enormous attention in recent years, due to its extensive applications in gaming, intelligent video surveillance, and human robot interaction. However, predicting the 3D human pose from the 2D image/pose is an ill-posed regression problem because of the inherent ambiguity between the 2D and 3D data. The inherent ambiguity and infinite possible of human poses lead to the requirement of large scale of well labeled 2D-3D pose pairs to train the estimation model. The main technologies of human 3D pose estimation can be categorized into two groups: the one-stage method~\citep{li20143d,pavlakos2017coarse,tekin2016structured} that directly regresses the human 3D pose from the 2D image and the two-stage method~\citep{drover2018can,fang2018learning,martinez2017simple,zhao2019semantic,zhou2019hemlets} that transits with the intermediate pose information extracted from the image. Directly regressing an image to the 3D human pose is a highly nonlinear problem, which leads to a large estimation model, a large solution searching space, and large chances to be sub-optimized. Therefore, lifting the predicted 2D human pose to the 3D pose springs up as one direction of the two-stage methods in this area. Many promising 2D pose lifting results~\citep{chen2019unsupervised,drover2018can,martinez2017simple,zhao2019semantic} have been reported. Estimating the 3D human pose from the 2D pose has many advantages: 1) the estimation of the 2D human pose from the image is a well-solved problem since learning methods~\citep{cao2017realtime,newell2016stacked,sun2019deep,wei2016convolutional} introduced into this area; 2) abundant 2D labels can be made use of, especially for in-the-wild scenarios. Labeling the 2D human pose is much easier than capturing the ground truth of the 3D human pose which requires expensive equipment and a controlled environment. 
In these regards, lifting the 2D human pose to the 3D human pose~\citep{nie2020unsupervised} and exploiting abundant 2D pose labels are two valuable topics that we focus on in this paper.

\begin{figure}[t]
    \begin{center}		
		\includegraphics[width=\linewidth]{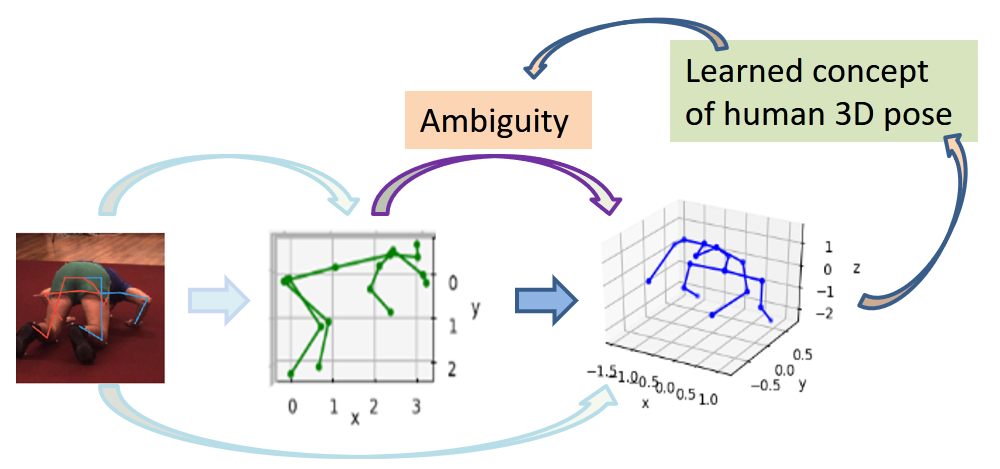}
	\end{center}
	\caption{The idea of this paper is to alleviate the ambiguity by 1) inherently modeling the 3D human pose and achieving a 3D concept of the human body 2) embedding the learned 3D body concept into the estimation process. Deep colored arrow illustrates our workflow.}
	\label{fig:idea}
\end{figure}

As discussed, the ambiguity between the 2D and 3D human poses is one of the biggest hindrances of achieving accurate 3D human pose estimation. Nonetheless, \cite{martinez2017simple} demonstrates that the 3D human pose can be attained with a remarkably low error rate from the 2D human pose through a simple linear neural network. Further, \cite{drover2018can} proved that the 3D human pose can be learned from 2D projections alone by randomly projecting the generated 3D skeleton back into the 2D space. Attributed to the prior knowledge of the human body that we have acquired in our mind, our human beings are able to efficiently recognize the 2D human pose and imaging its corresponding 3D human pose with the least ambiguity. Inspired by the perception manner of human beings, as shown in Fig.~\ref{fig:idea}, we will demonstrate that learning an inherent concept of the human body and embedding it into the 3D pose estimation process effectively alleviate the ambiguity, and improve the performance and the generalization of the 3D human pose estimation model. A well-learned body concept helps eliminate those impossible or unnatural solutions.

To learn a prior concept of the human 3D pose as embedding, we novelly treat the 2D human pose and the 3D human pose as two data domains. The 3D body concept is first learned from the 3D poses through an encoder-decoder architecture. To attain the 3D body concept consensus from 2D poses, we propose to use a domain adaptation~\citep{csurka2017domain,hoffman2018cycada,tzeng2017adversarial} architecture to map the 2D human pose and the 3D human pose into the same semantic space, as shown in Fig.~\ref{fig:leaning_embedding}. Through domain adaptation, the prior knowledge of the human body learned from 3D poses is transferred to the 2D pose encoder. The semantic representations learned from the 2D pose and 3D pose are further required to be able to reconstruct the corresponding 3D human pose through the same decoder, to make sure that they are informative enough. Actually, the representation of 2D pose can be deemed as a 3D "imagination" of the 2D pose based on the body concept that the network has learned.

Domain adaptation is proposed to solve the domain shift problem where the source data and target data have different distributions. By adapting the target data into the same hidden space of the source data, the knowledge learned from the source data can be transferred and applied to the target data. In previous works, domain adaptation is mainly applied on classification tasks~\citep{hoffman2018cycada,tzeng2017adversarial} or classification-related tasks, such as segmentation~\citep{li2019bidirectional}. We are the rare ones who utilize it to solve the regression problem. Integrating the domain adaptation architecture into the 3D pose regression model enables us to learn the manifold of the 3D human pose other than establishing a point-to-point regression model. More importantly, the proposed learning architecture based on domain adaptation can be easily extended to the semi-supervised mode, where the abundant 2D poses can be leveraged to improve the accuracy of 3D pose estimation. 

Our work not only achieves competitive SOTA results but also reveals the importance of the concept of human body in the performance and the generalization of a 2D pose lifting model. Although many previous works may utilize the body concept implicitly, we demonstrate that a well-learned and explicit 3D body concept can benefit the learning process most. In another aspect, compared to the semi-supervised classification (SSC), the semi-supervised regression (SSR) is an important but seldom explored topic~\citep{kostopoulos2018semi}. We hope our research brings inspiration to researchers not only in the human pose estimation community but also in the SSR community. 

Our main contributions are summarized as follows:
\begin{itemize}
	\item[$\bullet$] 
	We propose to learn a 3D body concept through the domain adaptation in regression to alleviate the 2D-3D ambiguity. Extensive experiments show that embedding the learned 3D body concept into the 2D human pose lifting process greatly improves the accuracy and generalization, especially when training data is scarce.
	\item[$\bullet$] 
	We present a new learning architecture for 3D pose estimation, which integrates supervised mode and semi-supervised mode in a unified manner without requiring any side information (such as multi views or temporal consistency) that many other weakly-supervised methods need.
	\item[$\bullet$] 
	Our framework significantly boosts the performance of the simple linear backbone network. State-of-the-art results are achieved in both the supervised mode and semi-supervised mode, which verifies that explicitly having a well-learned 3D body concept significantly benefits the 2D human pose lifting.  
\end{itemize}

\section{Related works}\label{sec2}

Lifting the 2D human pose to the 3D human pose is an important direction of the two-stage 3D human pose estimation.
In this section, approaches in this area are reviewed.

\subsection{2D human pose lifting}
Ambiguity between the 2D and 3D poses is the biggest hindrance for accurate pose estimation~\citep{fang2018learning,li2020cascaded,zhao2019semantic,wang2019not}.
\cite{martinez2017simple} demonstrated that the 3D human pose can be attained with a remarkably low error rate from the 2D human pose through a simple linear neural network. However, none of the knowledge of the human body is made use of to reduce the ambiguity. 
\cite{drover2018can} demonstrated that the 3D human pose can be learned from 2D projections alone. Their work exploits the prior of human 3D body implicitly by observing the reprojected 2D pose. 
\cite{fang2018learning} considered the relationship of different body parts and propose pose grammar for better 3D human pose recovery from 2D poses. 
\cite{zhao2019semantic} proposed a semantic graph model to encode the semantic relationship of body joints. 
\cite{wang2019not} argues that the importance of different joints are different and bi-directional dependencies of body parts are considered for better 3D human pose estimation. 
\cite{li2020cascaded} proposed a deep cascaded neural network to iteratively refine the predicted 3D poses. However, their improvement is mainly attained through a large scale of synthetic training poses. 
Different from these methods, we propose to learn a 3D concept of human body explicitly as embedding to reduce the inherent ambiguity, which improves the performance of the backbone network significantly, especially when data is scarce.

\subsection{Weakly or semi-supervised 3D human pose estimation}
Due to the infinite possibilities of the human pose, accurate 3D human pose estimation requires tons of well-labeled data, while achieving the 3D ground truth of human pose is high-cost in time and equipment. Therefore, recent methods~\citep{iqbal2020weakly,kocabas2019self,li2020cascaded,mitra2020multiview,rhodin2018learning,pavllo20193d} attempt to find more efficient weak supervision to train the model.
\cite{drover2018can} and \cite{chen2019unsupervised} proposed an unsupervised method by re-projecting the 3D human pose and find supervisions in the 2D space. 
Some other works built the weakly supervised 3D human pose lifting model by leveraging the multi-view consistency~\citep{iqbal2020weakly,kocabas2019self,mitra2020multiview,rhodin2018learning} or the time-consistency~\citep{pavllo20193d}. 
\cite{kundu2020self} developed a self-supervised method via synthesize novel images to reduce the data bias and improve the generalization. 
\cite{li2020cascaded} proposed a weakly-supervised method by generating novel 2D-3D human poses pairs.
Compared to previous weakly-supervised or semi-supervised models, the proposed method requires no side information, such as multi views, camera parameters, temporal consistency or synthetic data for semi-supervised learning. Our semi-supervised mode relies on generalizing the learned 3D body concept to unseen 2D pose and verifying the consistency of 3D body concept in the semantic space, as shown in Fig. 3.

\section{Our approach}\label{sec3}

\begin{figure}
    \begin{center}		
		\includegraphics[width=\linewidth]{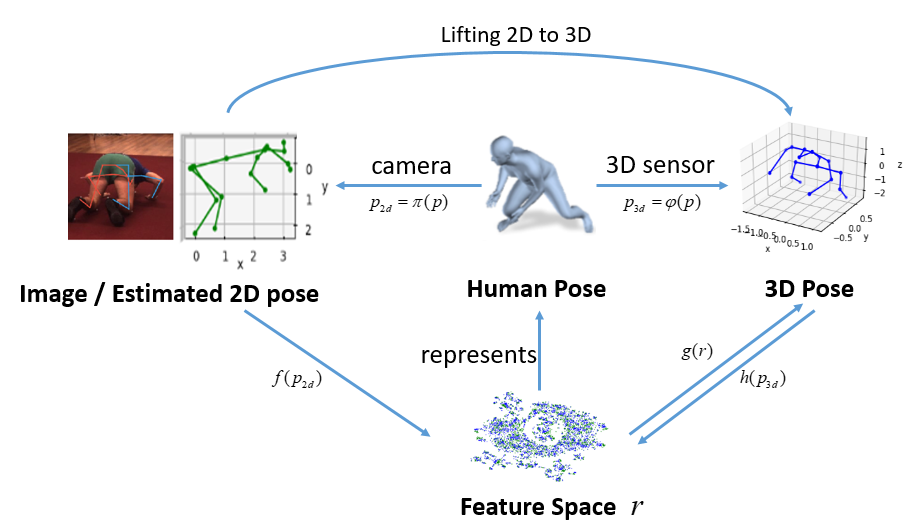}
	\end{center}
	\caption{Learning a 3D concept of the human body, where the 2D human pose and the 3D human pose are mapped into the same semantic space. The learned hidden representations are decoded back to the human 3D pose to avoid trivial solutions. }
	\label{fig:leaning_embedding}
\end{figure}

The ambiguity and absence of 3D ground truth in the wild scenarios are two of the biggest hindrances in achieving high accurate 3D human pose estimation from 2D poses. Human can image the corresponding 3D pose of a 2D human pose with the least ambiguity, which should be attributed to the common knowledge of human body we have acquired in our mind. Inspired by this, our idea is to alleviate the ambiguity by inherently modeling the 3D human pose and explicitly embedding the learned 3D prior of human body into the pose lifting network, as shown in Fig.~\ref{fig:idea}. The 3D prior of human body is learned through domain adaptation in the regression task. Learning the 3D body concept also enables us to extend the pose lifting framework to the semi-supervised mode where abundant 2D pose labels can be exploited to improve the 3D pose estimation accuracy. In the following, we will introduce our method from three aspects: learning the 3D concept of the human body through domain adaptation, 2D human pose lifting with the 3D "imagination" based on the 3D body concept, and extension to semi-supervised 3D human pose estimation.

\subsection{Learning the 3D body concept}
\begin{figure*}[t]
	\begin{center}		
		\includegraphics[width=0.9\linewidth]{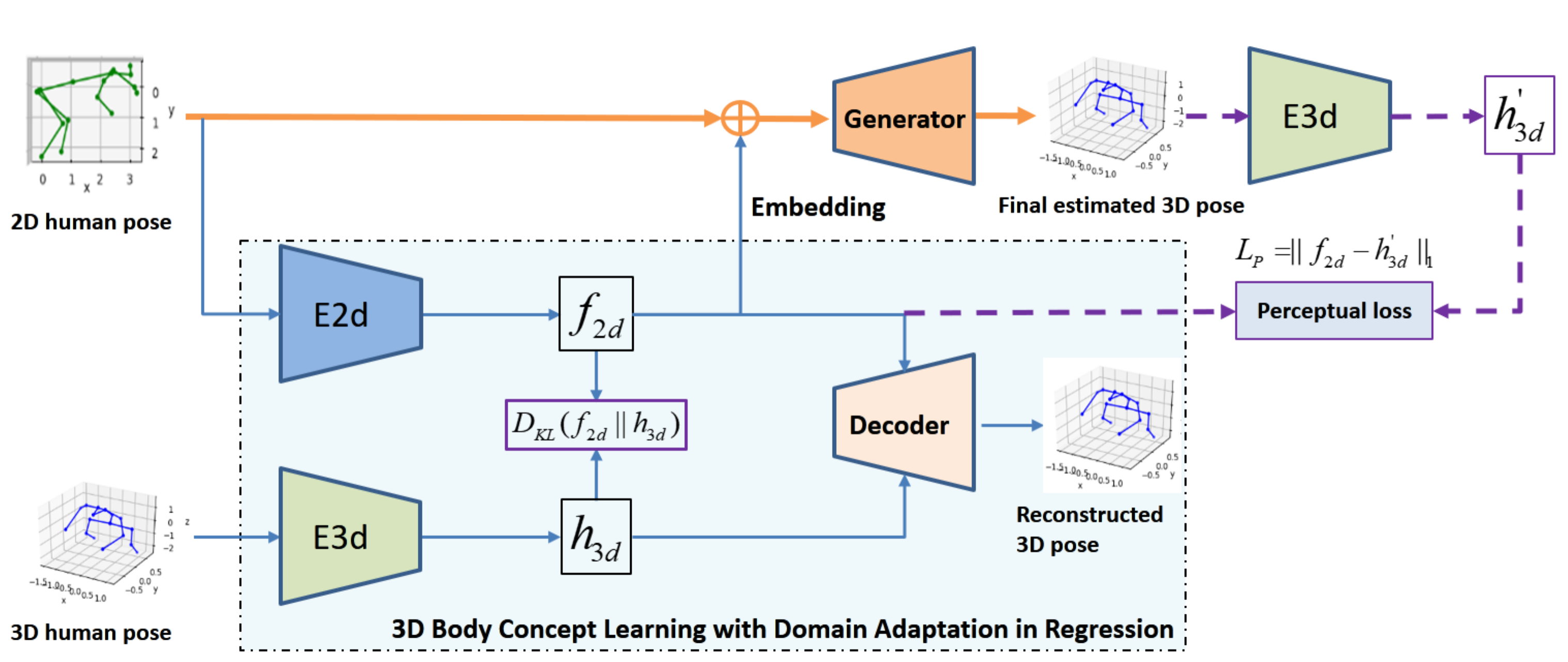}
	\end{center}
	\caption{3D human pose estimation with the 3D "imagination" of 2D pose as embedding. Modules in the light blue block is utilized for the 3D body concept learning, where the E2d and E3d denote the 2D pose encoder and 3D pose encoder. $f_{2d}$ and $h_{3d}$ are the semantic representation generated by encoders based on the body concept they have learned. The $f_{2d}$ which can be deemed as the 3D "imagination" of the 2D pose is further embedded into the main 2D pose lifting stream as colored in orange. The final predicted 3D pose is generated by the \textit{Generator}. In semi-supervised learning mode, there is no 3D pose reconstruction for unlabeled data but the predicted 3D poses are reprojected to the semantic space by reusing the 3D encoder as connected with dashed purple arrow.}
	\label{fig:full}	
\end{figure*}

Given a human pose, the 2d image/2d pose $p_{2d}$ and the 3D pose $p_{3d}$ can be obtained with the camera and 3D sensor respectively. We novelly treat the 2D human pose and the 3D human pose as two different data domains. Although these two domains have different distributions and unequal information, they, as descriptors of the same human pose, have inherent correspondence between each other. Therefore, the same body concept can be learned from the paired 2D and 3D human poses respectively. For unpaired 2D and 3D poses, the body concepts learned from them should be similar to each other, which is one of the bases of our semi-supervised pose estimation. Hence, to learn the 3D concept of the human body from paired or unpaired 2D-3D poses, a domain adaptation architecture is proposed for the pose regression task as shown in Fig.~\ref{fig:leaning_embedding}. The light blue block in Fig.~\ref{fig:full} shows the learning network corresponding to the Fig.~\ref{fig:leaning_embedding}. 

In Fig.~\ref{fig:leaning_embedding}, a basic ideal of autoencoder is utilized to learn a semantic representation $h(p_{3d})$ from the given 3D human pose in training data. Namely, the 3D pose is mapped into a hidden space and then decoded to reconstruct itself. If the learned semantic representation is non-trivial and be able to be decoded back to the 3D human pose, we believe that it encodes the essential feature of the human body and the 3D concept of the human body must have been acquired in the learned non-trivial semantic representation. To transfer the body concept learned from the 3D human pose to the encoding process of the 2D pose, the 2D human pose is mapped into the same semantic space as the 3D pose. Mathematically, the hidden representation of the 2D human pose $f(p_{2d})$ should have the same distribution with $h(p_{3d})$ and be able to generate the corresponding 3D human pose using the same pose decoder. That is, the 2D pose domain and 3D pose domain are adapted for the pose regression task as formulated in eq.~\ref{e1}.
\begin{equation}
 \label{e1}
\mathop{\arg\min}_{f, h, g} D_{KL}(P(f(p_{2d}))\|Q(h(p_{3d}))) + \|p_{3d} - g(r)\|_2
\end{equation}
\noindent where $D_{KL}$ is the Kullback-Leibler divergence between the distributions of the learned 2D representation $f(p_{2d})$ and the learned 3D representation $h(p_{3d})$. $f$ and $h$ denotes the functions of the 2D encoder and the 3D encoder respectively. $P$ and $Q$ are probability density functions. $g$ denotes the decoder function which infers the 3D pose from the semantic space. $r$ denotes the learned latent pose representation, which includes $f(p_{2d})$ and $h(p_{3d})$. Therefore, the reconstructed 3D pose from the 3D feature is $\hat{p}_{3d}=g(h(p_{3d}))$ and the estimated 3D pose from the 2D feature is $\hat{p}_{3d}^{\prime}=g(f(p_{2d}))$.

By adapting the 2D human pose into the same semantic space of the 3D human pose, the 3D concept of the human body learned from the 3D ground truth poses is transferred to the 2D encoder. Therewith, when fed with a 2D human pose, the 2D pose encoder actually images its corresponding 3D structure in the hidden space based on the 3D body concept it has acquired. In simple terms, we are using the concept learned from given 3D poses to teach the 2D encoder how to generate an intermediate informative 3D imagination from 2D poses for the 3D pose estimation. Benefits from such a learning manner, the manifold of 3D human pose is well learned other than establishing a point-to-point mapping model.

In practice, to avoid the 3D encoder learns a trivial hidden representation, the 2D encoder, 3D encoder and decoder are trained together. In this manner, the 3D encoder not only needs to generate an informative 3D representation but also should find a semantic space that the 2D mapping can easily reach. This is the reason why the proposed method can avoid trivial solution effectively.

\subsection{Lifting the 2D pose with 3D "imagination"}

The whole 3D pose lifting framework is shown in Fig.~\ref{fig:full}. It mainly contains two parts: the main 2D pose lifting stream with a pose generator to estimate the final 3D pose and a domain adaptation module to learn a 3D prior of the human body. Both the final 3D human pose lifting stream and the domain adaptation module need to predict the 3D human pose.

In previous section, given paired 2D and 3D poses $(p_{2d}, p_{3d})$, the domain adaptation module maps them into the same semantic space and generates 2D pose representation $f_{2d}$ and 3D pose representation $h_{3d}$ respectively. The target 3D human poses are reconstructed from both of the $f_{2d}$ and $h_{3d}$ through the same decoder. As $h_{3d}$ is learned from the 3D human pose, it encodes the inherent information of the 3D human body. Such a kind of knowledge will be transferred to the 2D encoder by regularizing $f_{2d}$ to have similar distribution with $h_{3d}$. That is to minimize the Kullback-Leibler divergence $D_{KL}$ between $f_{2d}$ and $h_{3d}$, which can be realized using the discriminator loss $L_D$ for unpaired data or the perceptual loss $L_P$ for pair data as shown in eq.~\ref{e2} and eq.~\ref{e3}, where $D$ is a domain discriminator which classifies whether the feature is generated from the 2D pose or the 3D pose. For paired $(p_{2d}, p_{3d})$, perceptual loss which directly constrains the generated semantic representation from 2D domain and 3D domain to be the same is stronger and preferred. But for unpaired 2D and 3D poses, discriminator loss can be utilized to require the obtained 2D representation $f_{2d}$ and 3D representation $h_{3d}$ are similar to each other. The 2D representation $f_{2d}$ is a hidden feature that contains some necessary 3D information for 2D pose lifting. Therefore, it can be treated as the 3D "imagination" of the 2D pose in hidden space, which are generated based on the concept of the human body learned from 3D poses.

\begin{align} 
\begin{split}\label{e2}
L_{D}=-\mathbb{E}_{p_{3d}}\left[ logD(h_{3d})\right] -          \mathbb{E}_{p_{2d}}\left[ log(1-D(f_{2d}))\right] \\
\end{split}\\
\begin{split}\label{e3}
L_P=\|f_{2d}-h_{3d}\|_1
\end{split}	
\end{align}
The generated 3D "imagination" of the 2D pose $f_{2d}$ is then embedded into the main 2D pose lifting stream to predict the final 3D human pose by concatenating it with the original 2D pose. Integrating all components together, the final loss function for the supervised human 3D pose estimation is shown as eq.~\ref{e4}.
\begin{equation} \label{e4}
L = \lambda_1 L_{est} + \lambda_2 L_P + \lambda_3 L_{rec}
\end{equation}
\noindent where the first part $L_{est}=\|p_{3d} - \hat{p}_{3d}^f\|_2$ is the final pose estimation loss of the generator, the second part $L_P$ is the perceptual loss in eq.~\ref{e3}, the third part $L_{rec}=\|p_{3d} - g(r)\|_2$ is the pose reconstruction loss in the domain adaptation module as defined in eq.~\ref{e1}. $\lambda_1$, $\lambda_2$, $\lambda_3$ are weights to balance the influence of the three loss. The whole architecture is trained end-to-end.

\noindent \textbf{Relationship between the pose generator and decoder:} Although the pose generator and decoder both have to predict the 3D pose corresponding to the 2D pose, their roles are different inherently. The decoder helps the 3D pose encoder learn the concept of the human body and leads the 2D encoder to generate useful 3D "imaginations" in the hidden space. It plays a role of assistant. While the generator is to predict the final 3D pose with the concatenation of the 2D pose and 3D "imagination" as input. Decoder roughly reconstructs the 3D pose while the generator estimates the 3D pose more finely. Therefore, the weight $\lambda_1$ for $L_{est}$ is ten times larger than the weight $\lambda_3$ for $L_{rec}$. The 3D "imagination" is a guidance for pose estimation task. For example, it is hard if we want to find something in our room without knowing any information. But if we figure out the shape of the target, things will be easier. The 3D "imagination" plays the role of telling pose generator what 3D human pose should be looks like based on the learned 3D body concept.

\subsection{Semi-supervised 3D human pose estimation}

Attributed to the learning of the 3D concept of the human body, the 3D pose estimation architecture can be efficiently extended to the semi-supervised learning mode by remapping the predicted 3D human pose into the semantic space using the trained 3D encoder, as shown in Fig.~\ref{fig:full}. The proposed semi-supervised pose estimation is based on two points: 1) the 3D body concept extract from the 2D human pose and the 3D human pose should be similar, i.e. $f_{2d}$ and $h_{3d}$ should have the same distribution; 2) the 2D pose and its predicted 3D pose (which is treated as fake label) should have self-consistency in the semantic space, i.e. $f_{2d}$ and the corresponding $h_{3d}^\prime$ should be the same. In other words, we extend the well-learned body concept from labeled 2D-3D pairs to unseen 2D human pose and its fake label.

Based on above two points, for unlabeled 2D human pose $p_{2d}^{ul}$, if the predicted 3D pose is correct, the remapped feature $h_{3d}^{ul}$ and $f_{2d}^{ul}$ should have self-consistency in the semantic space, as this is how the 2D encoder and 3D encoder are trained on labeled data in the domain adaptation module. Self-consistency in semantic space indirectly constrains the predicted 3D pose of the unlabeled 2D pose based on the 3D body concept the network has attained. Meanwhile, to make sure the 3D "imagination" are correctly generated from unlabeled 2D poses, $f_{2d}^{ul}$ is required to have a similar distribution with $h_{3d}^l$ of labeled 3D poses. Therefore, combining the training loss of the labeled data in eq.~\ref{e4} and loss for unlabeled data, the loss for semi-supervised 2D human pose lifting is formulated as eq.~\ref{e5}.
\begin{multline}\label{e5}
    L_{semi} = \lambda_1 L_{est}^l + \lambda_2 L_P^l(f_{2d}^{l}, h_{3d}^l) + \lambda_3 L_{rec}^l \\+ \lambda_4 L_D^{ul}(f_{2d}^{ul}, h_{3d}^l) + \lambda_5 L_P^{ul}(f_{2d}^{ul}, h_{3d}^{ul})
\end{multline} 

\noindent where the superscript $l$ denotes the labeled data and $ul$ denotes the unlabeled data. $\lambda_4$ and $\lambda_5$ are weights to adjust the influence of the unlabeled data. $\lambda_4$ and $\lambda_5$ are smaller than the weights for labeled data. $L_{est}$, $L_P$, $L_D$, and $L_{rec}$ are defined as previous section. Although the perceptual loss is applied on both of the labeled 2D poses and unlabeled 2D poses, it should be noted that their calculations are different. Perceptual loss $L_P^l$ is calculated between the 2D pose and its true 3D pose from the labeled data. While the perceptual loss $L_P^{ul}$ defines the discrepancy between the unlabeled 2D pose and its fake label (predicted 3D poses) to seek self-consistency in the semantic space. $L_{est}$ and $L_{rec}$ are not applied to unlabeled data.

It's also noteworthy that the proposed semi-supervised architecture doesn't require any side information, such as temporal information~\citep{pavllo20193d}, camera information for re-projecting to 2D space~\citep{habibie2019wild}, multi-view information~\citep{kocabas2019self} or synthetic training data~\citep{drover2018can, li2020cascaded} that previous weakly-supervised methods commonly need. In contrast, we are seeking for a consistency of body concept on the 2D human pose, 3D human pose and the predicted 3D human pose. Moreover, the proposed semi-supervised architecture adds no more parameters to the supervised mode by reusing the 3D encoder. In these regards, our method is more efficient. Owning to no requirement for specific information, the proposed learning architecture has potential to be applied to other regression problem.

\section{Experiments}
\label{experiments}
In the following, we evaluate the performance of our learning architecture and verify the effectiveness of 3D body concept embedding in reducing the ambiguity and improving the generalization.

\subsection{Datasets and evaluation metrics}
\textbf{H3.6M} which contains millions of human pose is the largest benchmark for the 3D human pose estimation. These poses come from 15 different motions performed by 11 professional actors. According to~\cite{h36m_pami}, data of the subjects 1, 5, 6, 7, 8 is utilized for training. Poses of subjects 9 and 11 are left for testing. The Mean Per Joint Position Error (MPJPE, mm), which refers to the average Euclidean distance between the estimated joints and the ground truths, is adopted as the metric of the model performance. 

Usually two protocols are used in the evaluation. Protocol \#1 only aligns the root joints of the ground truth pose and the estimated 3D pose before calculating the MPJPE. Protocol \#2 aligns the 3D prediction with the ground truth via a rigid transformation and scaling. Besides the two cross-subject evaluation protocols, we propose a cross-action protocol \#3 to evaluate the generalization of trained models. In Protocol \#3, poses of several actions are used for training and poses of the other actions are left for testing. Thus, a big discrepancy exists between the training pose and testing pose. This protocol requires the model to make inference on unseen poses, which tests whether the 3D body concept are well-learned by our model. 

Testing of a method can be implemented on the ground truth (GT) 2D poses that are projected from the GT 3D poses and detected 2D poses estimated from images. Evaluation on the GT 2D poses shows the performance upper bound of a method. Namely, with improvement of the 2D human pose detection, how good a method can be in lifting the 2D human pose.

\textbf{MPI-INF-3DHP} is mainly used for the generalization evaluation. Different from H3.6M that only contains indoor scenes, it includes both the indoor and outdoor poses. Model pretrained on the H3.6M dataset is applied on the test data of MPI-INF-3DHP without finetune. Besides MPJPE, Percentage of Correct Keypoints (PCK) which measures the correctness of 3D joint predictions under a specified threshold and Area Under the Curve (AUC) computed for a range of PCK threshold are also reported.

\subsection{Backbone network}
\label{backbone}
\begin{figure}[t]
    \vspace{6pt}
	\centering
	\includegraphics[width=\linewidth]{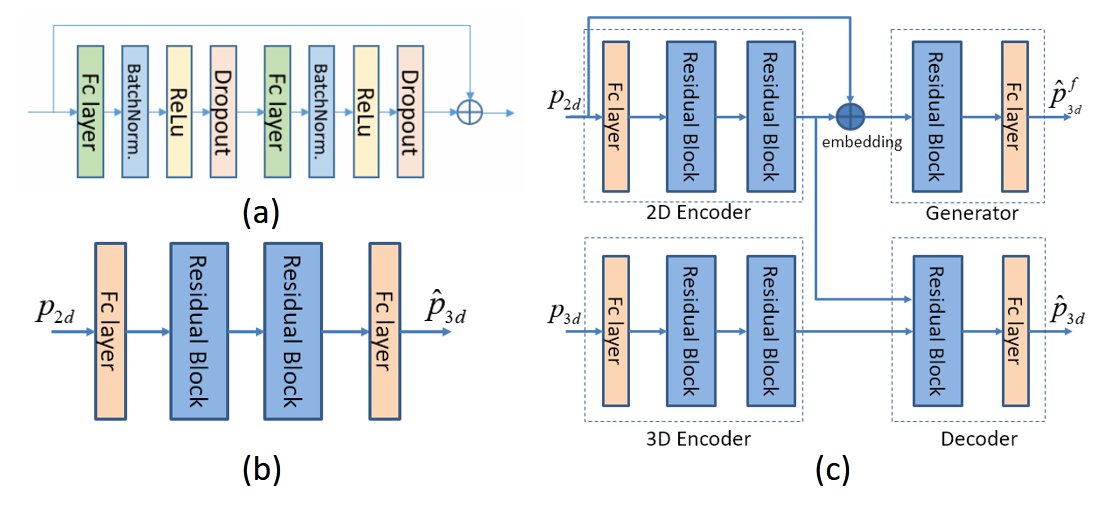}
	\caption{Details of networks: (a) details of the linear residual block, which is used as the basic module, (b) the linear network proposed by \cite{martinez2017simple}, (c) the network used in our learning architecture, which is based on the linear network.}		
	\label{fig:backbone}       
\end{figure}

To demonstrate the effectiveness of the proposed learning architecture, the simple linear network proposed by \cite{martinez2017simple} is used as the backbone network to establish the encoders, decoder and pose generator. The basic module of this linear network is a linear residual block as shown in Fig.~\ref{fig:backbone}(a). The linear residual block consists of two fully connected layers, batch normalization layers, ReLU activation layers, and dropout layers. Attributed to the effectiveness of the linear residual block, it is widely adopted in many 2D pose lifting networks~\citep{pavllo20193d,fang2018learning,li2020cascaded}. Following~\cite{martinez2017simple}, we set the dropout rate to 0.5. Our 2D and 3D encoder have the same structure which consists of two residual blocks and a fully connected layer before the residual blocks. The decoder and pose generator have the same structure which is composed of a residual block and a fully connected layer at the output end.

\subsection{Implementation details}
\label{implementation_details}
The simple linear network proposed by \cite{martinez2017simple} is used as the backbone network to establish the encoders, decoder and pose generator. Adam optimizer with an initial learning rate of 0.001 is utilized in training. Weights to adjust the influence of unlabeled data defined in eq.~\ref{e5} are set to $\lambda_1 = 10, \lambda_2 =1, \lambda_3 = 1, \lambda_4 =0.1, \lambda_5 = 0.5$. The batch size is 64. When using the discriminator loss, the discriminator is simply consisted of 3 fully connected layers with output size of 512, 1024, and 1. ReLU activation is used after each fully connected layer except the output layer. Joint of nose in the 2D human pose is not used when comparing with \cite{martinez2017simple} and is used when comparing with \cite{li2020cascaded}, just follow these works. Our experiments are implemented on a computer with 2 NVIDIA Titan XP graphics cards, 64 GB RAM, and an Intel Xeon(R) processor E5-2640.

\subsection{Ablation study}
\noindent\textbf{Effect of the 3D Body Concept.}
Our pose generator, encoders and decoder are built using the same linear network in~\cite{martinez2017simple} which is consisted of several linear residual blocks as shown in Sec.~\ref{backbone}. Therefore, the difference of the performance between our method and \cite{martinez2017simple} can be attributed to the embedding of the 3D body concept in 2D pose lifting.

\begin{figure*}[t]
	\centering
	\includegraphics[width=0.85\linewidth]{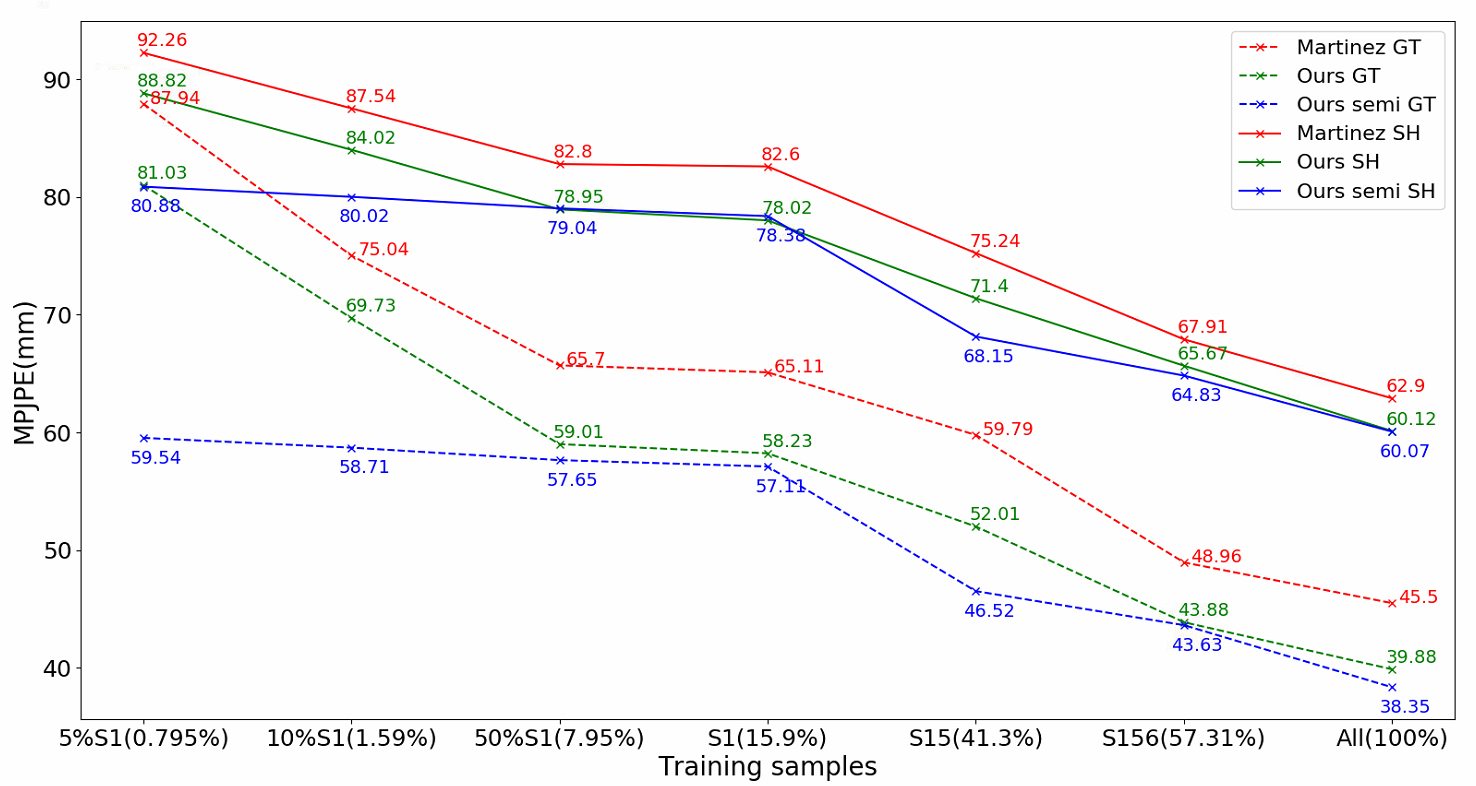}
	\caption{Comparison of performance between our method and \cite{martinez2017simple} under protocol \#1, where the "GT" means using ground truth 2D human poses for training and "SH" means using the detected 2D human poses with Stacked Hourglass method~\citep{newell2016stacked} as ~\cite{martinez2017simple} did.}		
	\label{fig:compare}       
\end{figure*}

Performance comparison between our method and \cite{martinez2017simple} is shown in Fig.~\ref{fig:compare}, where the horizontal axis represents the size of training data. "S1" means the poses of subject 1. "S15" denotes the pose collection of subject 1 and 5. "All" refers to the whole training set in H3.6M, i.e., S15678. The percentage in parenthesis shows the proportion of the leveraged training data to the whole training set. Both the results of full supervised mode and semi-supervised mode are reported. When training in semi-supervised mode, besides the labeled data utilized in the corresponding full supervised mode, the rest data in the whole training set is utilized as unlabeled data. To have a fair comparison, we use the same GT 2D poses and detected 2D human poses achieved with Stacked Hourglass (SH) method~\citep{newell2016stacked} as \cite{martinez2017simple}.

Fig.~\ref{fig:compare} clearly shows that our method performs better than \cite{martinez2017simple} by a large margin, either in full supervised mode or semi-supervised mode. Such improvement brought by embedding the 3D body concept into the lifting network is consistent when evaluated on the GT 2D poses and noisy detected 2D poses. In particular, when the data is scarce, the promotion of accuracy is much more salient. For example, with the 3D body concept, when training with only 5\%S1 of the SH 2D poses as labeled data, the performance of~\cite{martinez2017simple} is improved by 3.73\% in supervised mode and 12.33\% in semi-supervised mode. Using the same size of GT 2D human poses, the increment of accuracy are 7.86\% in supervised mode and 32.44\% in semi-supervised mode. That is to say, with only 0.795\% of the training data, the proposed method attains a 77.8\% performance of \cite{martinez2017simple} that uses the whole training set. Even when all training data are used, our method still attains an improvement of 4.42\% on SH 2D human pose and 12.4\% on the GT 2D human pose in supervised mode. These results demonstrate that the well-learned 3D concept of human body from 3D poses boosts the accuracy of 3D pose estimation.

\begin{table*}[ht]
	\begin{center}
	\caption{Detail contrast between our method and \cite{martinez2017simple}, where "sup." denotes implementation in supervised mode and "semi." denotes semi-supervised mode. "$\uparrow$" means the performance is increased and "$\downarrow$" means decrease of the accuracy. It shows our method performs better with a large margin both in supervised mode and semi-supervised mode.}\label{tab:improvement}
	\resizebox{\linewidth}{!}{
		\begin{tabular}{l|l|c|c|c|c|c|c|c|c}
			\hline
			Data &Method & 5\%S1 & 10\%S1 & 50\%S1 &S1 &S15 &S156 & All & Avg.\\
			\hline\hline
			\multirow{3}{*}{SH} &Ours sup.$-$\cite{martinez2017simple} & ${\uparrow 3.73\%}$ & ${\uparrow 4.02\%}$ & ${\uparrow 4.65\%}$ & ${\uparrow 5.54\%}$ & ${\uparrow 5.10\%}$ & ${\uparrow 3.30\%}$ & ${\uparrow 4.42\%}$ & ${\uparrow 4.4\%}$ \\
			&Ours semi.$-$\cite{martinez2017simple}& ${\uparrow 12.3\%}$ & ${\uparrow 8.59\%}$ & ${\uparrow 4.54\%}$ & ${\uparrow 5.11\%}$ & ${\uparrow 9.42\%}$ & ${\uparrow 4.54\%}$ & ${\uparrow 4.50\%}$ & ${\uparrow 7.0\%}$\\
			&Ours semi.$-$Ours sup. & ${\uparrow 8.61\%}$ & ${\uparrow 4.57\%}$ & ${\downarrow 0.11\%}$ & ${\downarrow 0.44\%}$ & ${\uparrow 4.32\%}$ & ${\uparrow 1.24\%}$ & ${\uparrow 0.08\%}$ &${\uparrow 2.6\%}$\\
			\hline
			\multirow{3}{*}{GT} &Ours sup.$-$\cite{martinez2017simple} & ${\uparrow 7.86\%}$ & ${\uparrow 7.08\%}$ & ${\uparrow 10.2\%}$ & ${\uparrow 10.6\%}$ & ${\uparrow 13.0\%}$ & ${\uparrow 10.4\%}$ & ${\uparrow 12.4\%}$  & ${\uparrow 10.2\%}$\\
			&Ours semi.$-$\cite{martinez2017simple}& ${\uparrow 32.3\%}$ & ${\uparrow 21.8\%}$ & ${\uparrow 12.3\%}$ & ${\uparrow 12.3\%}$ & ${\uparrow 22.2\%}$ & ${\uparrow 10.9\%}$ & ${\uparrow 14.3\%}$  & ${\uparrow 18.2\%}$\\
			&Ours semi.$-$Ours sup.& ${\uparrow 24.4\%}$ & ${\uparrow 14.7\%}$ & ${\uparrow 2.07\%}$ & ${\uparrow 1.72\%}$ & ${\uparrow 9.18\%}$ & ${\uparrow 0.51\%}$ & ${\uparrow 3.36\%}$  & ${\uparrow 8.0\%}$\\
			\hline
		\end{tabular}}
	\end{center}
\end{table*}

Detail increments brought by the proposed method is shown in Table~\ref{tab:improvement} corresponding to Fig.~\ref{fig:compare}. Averagely, in supervised mode, embedding the 3D imagination into estimation performs 4.4\% better than the method~\citep{martinez2017simple} that lifts 2D human poses directly. When evaluated on clear GT 2D human pose, the upper bound performance is promoted by 10.2\% on average. The results demonstrate our idea that having a well-learned 3D concept of human body helps alleviate the ambiguity of lifting the 2D human pose. Such a kind of improvement is consistent when increasing the size of training data, as shown in Fig.~\ref{fig:compare}. At most of the time, the semi-supervised mode performs better than the supervised method, which supports another motivation of our work to leverage the abundant unlabeled 2D poses.

\noindent\textbf{Robustness over different 2D Pose Sources.}
\label{sec:2deffects}
Performance on the GT 2D human pose reveals the upper bound performance of our method. To further demonstrate the effectiveness and robustness of our method on real detected 2D human poses, the proposed method is compared with one of the most recent works of \cite{li2020cascaded} which utilized a 2D detector HR-Net~\citep{sun2019deep}. Table~\ref{tab:cascaded} shows the results on different 2D poses generated by different methods. Different from our method which embeds the 3D body concept into the 2D pose lifting process, Li~et al. proposed a cascaded network to iteratively use the predicted 3D human pose for refinement. Compared to the baseline result of \cite{martinez2017simple} on GT 2D poses, the cascaded deep network~\citep{li2020cascaded} doesn't improve the accuracy too much. Their performance is mainly acquired by synthesizing additional training data. While our method attains more significant improvement either with or without synthetic training data. On the detected 2D poses of HRN method, our method still performs better than \cite{li2020cascaded}, which shows the robustness of our method in handling 2D poses with different quality. In fact, from the performance on SH, HRN and GT 2D poses we can see that, \textit{with improving the 2D pose detection, the proposed method could perform better}. Comparison with \cite{li2020cascaded} also indicates that explicitly learning a good 3D concept of the human body from 3D poses is more effective than only utilizing the 3D pose as supervision for 3D pose estimation. 

\begin{table}
    \caption{Results on 2D poses generated by different methods on H3.6M. The "E" represents the additional training data that generated by \cite{li2020cascaded}. Evaluation is implemented under protocol \#1.}
	\label{tab:cascaded}
	\centering
    \resizebox{\linewidth}{!}{
	\begin{tabular}{l|l|c|c|c|c}
		\toprule
		2D Source &Method &S1 &S1 + E & All & All + E\\
		\hline\hline
		\multirow{3}{*}{SH}
		&\cite{martinez2017simple} &82.8 &- &62.9 &- \\
		&Ours sup. &78.02 &- &60.12 &-\\
		&Ours semi. &78.38 &- &60.07 &-\\
		\hline
		\multirow{3}{*}{HRN} 
		&\cite{li2020cascaded} &70.1 &62.9 &52.1 &50.9 \\
		&Ours sup. &64.3 &59.9 &51.2 &50.7\\
		&Ours semi. &65.1 &59.5 &51.1 &50.3\\
		\hline
		\multirow{4}{*}{GT} &\cite{martinez2017simple} & 65.1 & - & 45.5 & -\\
		&\cite{li2020cascaded} &64.5 &50.5 &42.9 &34.5 \\
		&Ours sup. &56.9 &48.0 &39.9 &33.7\\
		&Ours semi. &55.4 &47.1 &38.0 &35.5\\
		\hline
	\end{tabular}}
\end{table}

\begin{figure}
    \centering
    \includegraphics[width=\linewidth]{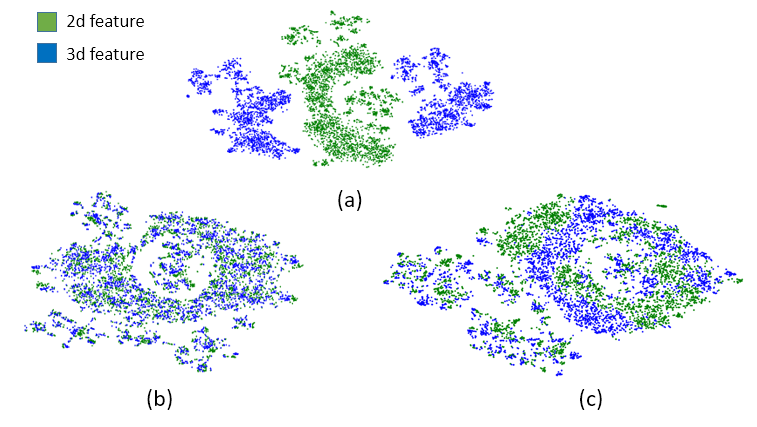}
    \caption{Distribution of the learned semantic 2D and 3D features visualized by t-SNE analysis: (a) the 2D and 3D stream are trained independently (b) the pose domain adaptation with perceptual loss (c) the pose domain adaptation with discriminator loss}
    \label{fig:hiddenfeature} 
\end{figure}

\noindent\textbf{Effect of the domain adaptation in regression.}
The domain adaptation module is proposed to learn the 3D concept of human body using the 3D poses in labeled training data. To figure out how the domain adaptation module works in learning the 3D concept of human pose, the learned semantic features from 2D pose and 3D pose are visualized in Fig.~\ref{fig:hiddenfeature} using t-SNE analysis. In Fig.~\ref{fig:hiddenfeature}, (a) shows the distribution of 2D and 3D features generated by training the 3D stream and the 2D stream independently. It should be noted that training the 2D stream independently equals to the method of \cite{martinez2017simple}. As can be seen, the feature learned from 2D poses has a completely different distribution with the feature learned in the 3D stream, although the two streams have to reconstruct the same 3D poses. Such a difference indicates that the 2D pose is lifted in a different manner with the pose reconstruction in the 3D stream. \textit{As a consequence, 2D pose lifting with only a 3D supervision at the output end may not truly capture the intrinsic feature of human 3D pose, which affects the efficiency and generalization of those methods based on it.}

However, with the domain adaptation in regression, as shown in (b) and (c), the 2D feature and 3D feature are located in the same space, which means they encodes the same feature of human body. The domain adaptation module enforces the 2D stream and 3D stream to learn the same semantic understanding of the human body that benefits the reconstruction of 3D human poses. As the learned semantic understanding first is a latent representation of the 3D pose, it encodes the 3D concept of human body in hidden space. The learned concept of human body will be transferred to the 2D encoder during domain adaptation and guides the 2D pose encoder to generate informative 3D "imagination" of the 2D pose. Knowing what a human 3D pose should be helps alleviate the ambiguity in lifting the 2D human pose to 3D, as well as improving the generalization of the trained model. 

\subsection{Comparison on supervised 3D pose estimation}
Comparison between the proposed method and other state-of-the-art supervised methods is shown in Table~\ref{tab:supCom}. It can be seen that our method achieves quite competitive results to those SOTA methods. Note that methods marked with superscribe \textbf{\&} integrates additional information for predicting the 3D pose, such as multi views~\citep{iqbal2020weakly,kocabas2019self,mitra2020multiview}, temporal consistency~\citep{pavllo20193d}, manual defined relationship between different body parts~\citep{wang2019not}, ordinal information~\citep{sharma2019monocular}. However, as \cite{martinez2017simple} and \cite{zhao2019semantic}, our method lifts the detected 2D key-points to the 3D human pose with none of the above side information. Hence, results of our method show the pure capability of the proposed learning architecture in lifting 2D human pose. In fact, without using the temporal information, results of \cite{pavllo20193d} which are the best in Table~\ref{tab:supCom} decrease to 51.8 under P\#1 and 40.0 under P\#2, both are inferior to ours. By learning a 3D concept of the human body, our network implicitly learns the relationship of different body parts. Our method achieves the best results among those methods without additional side information. In Table~\ref{tab:supCom}, by using the test data as unlabeled data, the inference result of semi-supervised mode is also reported. To efficiently leverage unlabeled 2D poses is the advantage of our method. 

\begin{table}[t]
    \begin{center}
    \caption{Comparison between our method and other SOTA full supervised methods on H3.6M dataset. All training data is used. P\#1 and P\#2 are the protocols used. P\#1$^*$ is P\#1 evaluated on GT 2D poses. Methods with \& means additional information is utilized, such as multi views or temporal consistency.}
	\label{tab:supCom}
	\resizebox{\linewidth}{!}{
		\begin{tabular}{l|c|c|c}
			\toprule
			\multirow{2}{*}{Method} & \multicolumn{3}{c}{MPJPE(mm)}\\
			\cline{2-4}
			&P\#1 &P\#1* & P\#2 \\
			\hline\hline
			\cite{mitra2020multiview}$^\&$(CVPR)~ & 94.25 &- &72.48 \\
			RepNet(CVPR)\\~\citep{wandt2019repnet} &89.9 &50.9 &65.1 \\
			\cite{martinez2017simple}(ICCV) &62.9 & 45.5 & 47.7 \\
			\cite{sharma2019monocular}(ICCV)   & 58.0 &- &40.9 \\
			\cite{zhao2019semantic}(CVPR) & 57.6 &43.8 &-  \\
			\cite{iqbal2020weakly}$^\&$(CVPR) & 56.1 & - & 45.9\\
			\cite{moon2019camera}(ICCV) & 54.4 &35.2 & - \\
			\cite{wang2019not}(ICCV) & 52.6 & - &40.7 \\
			\cite{kocabas2019self}$^\&$(CVPR) &51.83 &- &45.04 \\
			\cite{pavllo20193d}$^\&$(CVPR) &\textbf{46.8} &- &\textbf{36.5} \\
			\cite{li2020cascaded}(CVPR) &50.9 &34.5 &38.0 \\
			\textbf{Ours} &50.7 &\textbf{33.68} &37.7 \\
			\textbf{Ours semi.}(transductive)  &50.3 &35.48 &37.3 \\
			\bottomrule
		\end{tabular}}
	\end{center}
\end{table}

\subsection{Comparison on semi-supervised 3D pose estimation}
\label{semi_com}
Besides ambiguity, lack of well labeled 2D-3D human pose pairs is another big challenge for the real application. Our learning framework provides an efficient approach to leverage the abundant unlabeled 2D human poses. As have been seen in Fig.~\ref{fig:compare}, and Table~\ref{tab:cascaded}, exploiting those unlabeled 2D poses is able to improve the model's performance, especially when the training data is scarce. With the increment of labeled data, the effect of unlabeled data decreases. It is noteworthy that the benefit brought by the unlabeled data also depends on its quality or the discrepancy between unlabeled data and test data, which is a more complicated topic we are not going to discuss in this paper. Despite the quality of unlabeled data, results of our experiments demonstrate the general effectiveness of the proposed semi-supervised method in leveraging unlabeled data.

\begin{table}
\begin{center}
    \caption{Comparison of our method and other SOTA semi-supervised or weakly-supervised methods. Only data of S1 in H3.6M is used as labeled data. P\#1, P\#1$^*$ and P\#2 are the same with Table~\ref{tab:supCom}. Except the method marked with "${All}$" is trained with the whole training set, other results are attained only using S1 as labeled data.}
	\label{tab:semiCom}
	\resizebox{\linewidth}{!}{
		\begin{tabular}{l|c|c|c}
		\toprule
			\multirow{2}{*}{Method} & \multicolumn{3}{c}{MPJPE(mm)}\\
			\cline{2-4}
			&P\#1 &P\#1* & P\#2 \\
			\hline\hline
			\cite{mitra2020multiview}$^\&$(CVPR) & 120.95 &- &90.76 \\
			\cite{li2019boosting}$^\&$(ICCV) &88.77 &- &66.5 \\
			\cite{kocabas2019self}$^\&$(CVPR) &65.35 &- &57.22 \\
			\cite{kundu2020self}(CVPR) & - & - & 50.8 \\
			\cite{pavllo20193d}$^\&$(CVPR) &64.7 &- &- \\
			\cite{rhodin2018learning}$^\&$(CVPR) &- &- &64.6 \\
			\cite{drover2018can}$^{All}$(ECCV) & - & - & 64.6 \\
			\cite{li2020cascaded}(CVPR) &62.9 &50.5 &47.5 \\
			\cite{iqbal2020weakly}(CVPR) & 59.7 & - & 50.6\\
			\cite{gong2021poseaug}(CVPR) & - & 56.7 & -\\
			\textbf{Ours semi.}(inductive) &\textbf{59.5} &\textbf{47.1} &\textbf{43.25} \\
		\bottomrule
		\end{tabular}}
	\end{center}
\end{table}

\noindent\textbf{Transductive Inference.} The significance of semi-supervised learning is that the unlabeled data including the target testing data can be used to improve the performance and adapt the trained model to the target scenarios. In Fig.~\ref{fig:compare}, Table~\ref{tab:cascaded}, and Table~\ref{tab:supCom}, using "All" training data in semi-supervised mode is the transductive inference where the testing data is used as the unlabeled data. Even when the size of training data is large, such a kind of transductive inference is able to further increase the accuracy of our supervised method by 0.08\%, 0.09\% and 4.76\% on SH, HRN, and GT 2D poses respectively.

\noindent\textbf{Inductive Inference.} In this case, the test data is not used as part of the unlabeled data in training. As can be seen from the reported results in Table~\ref{tab:improvement}, our method shows a good performance in the inductive inference. For example, when training with only 0.8\% (5\%S1) of the detected SH 2D poses, the accuracy is increased by 12.3\%  and 8.61\%  respectively compared to the result of \cite{martinez2017simple} and the result of our method without exploiting the unlabeled 2D poses. On detected HRN 2D poses, as shown in Table~\ref{tab:cascaded}, the increment brought by unlabeled poses is 8.3\% contrasted to the deep cascaded network~\citep{li2020cascaded}. 

Comparison between our method in the semi-supervised mode and other SOTA semi-supervised or weakly-supervised methods is shown in Table.~\ref{tab:semiCom}, where only poses of S1 in H3.6M dataset are used as labeled data for training. It shows that our method achieves the best results under both P\#1 and P\#2.

Note that our semi-supervised method doesn't require any side information, such as camera information to re-project the 3D poses, multi-view information~\citep{iqbal2020weakly,kocabas2019self,mitra2020multiview,rhodin2018learning}, and temporal information~\citep{li2019boosting,pavllo20193d} used by other methods. The proposed semi-supervised method purely attempts to generalize the intrinsic understanding of the human pose it has learned to those unlabeled 2D poses. Performing better than methods with additional information reveals the importance and effectiveness of having a well-learned 3D concept of the human body to the accurate 3D human pose estimation and model generalization, particularly when the well-labeled data is rare. We also argue that the proposed method can perform better if combined with the additional information.

\subsection{Generalization evaluation}

\begin{table}
    \caption{Cross-action 2D pose lifting on the H3.6M dataset. The MPJPE is calculated under P\#1.}
	\label{tab:crossaction}
	\begin{center}
        \resizebox{\linewidth}{!}{
		\begin{tabular}{l|l|c|c|c}
			\toprule
			\multicolumn{2}{c|}{Data} &\cite{martinez2017simple} &Ours & Ours semi.\\
			\hline
			\multirow{2}{*}{A1} 
			&SH &135.4 &124.9 &123.7 \\
			&GT &126.8 &121.8 &112.5 \\
			\hline\hline
			\multicolumn{2}{c|}{ } &\cite{li2019boosting} & Ours & Ours semi. \\
			\hline
			{A1-8} 
			&SH &88.75 &\textbf{75.56} &79.79 \\
			\bottomrule
		\end{tabular}}
	\end{center}
\end{table}

\begin{table}
    \caption{Cross-dataset evaluation on MPI-INF-3DHP dataset. Annotated 2D keypoints are used and models are pre-trained on H3.6M dataset. PCK and AUC are the higher the better.}
    \vspace{-6pt}
    \label{tab:crossdataset}
    \begin{center}
    \resizebox{\linewidth}{!}{
    \begin{tabular}{l|c|c|c}
    \toprule
         Method & PCK & AUC & MPJPE\\
         \hline
         \cite{zhou2017towards} & 69.2 & 32.5 & 137.1\\
         \cite{martinez2017simple} &75.3 &35.1 &105.8 \\
         \cite{mehta2017monocular} &76.5 &40.8 &117.6 \\
         \cite{mehta2018single} & 75.2 & 37.8 & 122.2 \\
         \cite{kanazawa2018end} &77.1 &40.7 &113.2\\
         \cite{yang20183d} &69.0 &32.0 &- \\
         \cite{luo2018orinet} &64.6 &32.1 &-\\
         \cite{li2019generating} &67.9 &- &- \\
         \cite{li2020cascaded} &81.2 &46.1 &99.7 \\
         \cite{wandt2019repnet} &81.8 &\textbf{54.8} &92.5 \\
         \hline
         \textbf{Ours} & 83.2 & 44.9 & 85.71 \\
         \textbf{Ours semi.} (inductive) &\textbf{83.5} &45.9 &\textbf{83.03}\\
    \bottomrule
    \end{tabular}}
    \end{center}
\end{table}

To evaluate the generalization, we test the proposed method using an additional cross-action protocol P\#3 and an unseen dataset MPI-INF-3DHP under the cross-dataset setting. The cross-action protocol trains the model on poses from a few actions and tests it on poses from the other unseen actions. The cross-action evaluation results are shown in Table~\ref{tab:crossaction}. In Table~\ref{tab:crossaction}, the selected training action "A1" is "Directions". The testing data are the full set of S9 and S11, which includes 14 unseen actions like "sitting", "phoning", "smoking", etc. To compare with \cite{li2019boosting}, "A1-8" means poses of the first 8 actions of training set defined in H3.6M are used for training and poses from the rest 7 actions are used for testing. 

The cross-dataset setting applies the model which is pretrained on the H3.6M dataset to the unseen test data in MPI-INF-3DHP without any training. As~\cite{kanazawa2018end,li2020cascaded,li2019boosting,luo2018orinet}, we report all of the PCK, AUC, and MPJPE in the cross-dataset evaluation in Table~\ref{tab:crossdataset}.

Table~\ref{tab:crossaction} verifies the good generalization of our method in processing unseen poses from different actions. Table~\ref{tab:crossdataset} shows the capability of our method in handling unseen data from different scenarios. The excellent generalization under either the cross-action or the cross-dataset setting is brought by the well-learned inherent 3D body concept, which is notably salient compared to the baseline performance of \cite{martinez2017simple}.

\begin{figure*}[ht]
    \vspace{12pt}
	\centering
	\includegraphics[width=\linewidth]{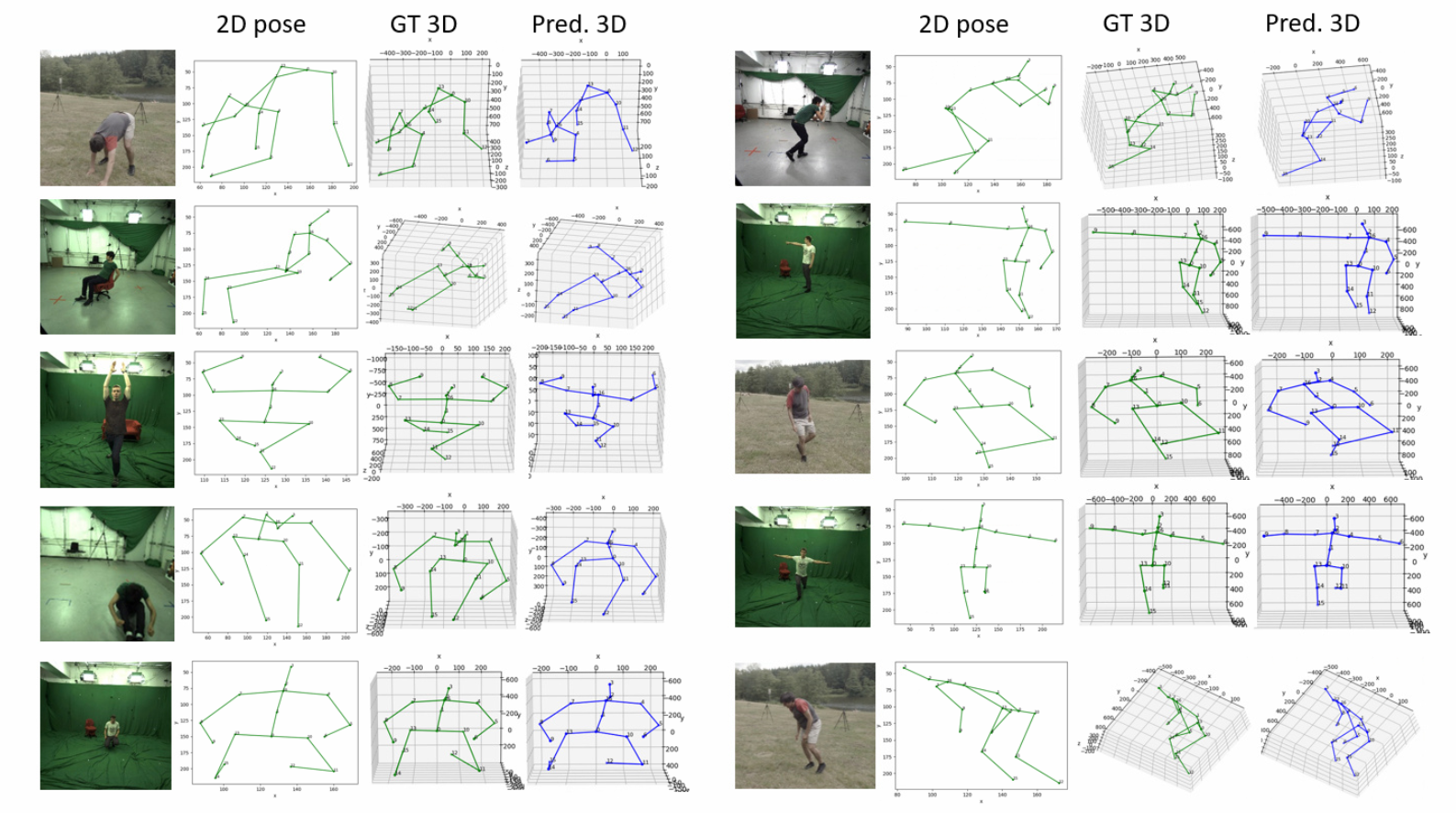}
	\caption{Qualitative evaluation under cross-dataset setting on MPI-INF-3DHP dataset. The 3D poses are rotated to have a better 3D view. Predicted poses are show in blue. }		
	\label{fig:qualitative_eval}       
\end{figure*}

\subsection{Qualitative evaluation}
Qualitative evaluation on MPI-INF-3DHP dataset which includes both indoor and outdoor scenarios is show in Fig.~\ref{fig:qualitative_eval}. The input 2D poses and GT 3D poses are drawn in green. The predicted 3D poses are drawn in blue. Rigid alignment is used. It can be seen that, the predicted 3D poses well aligns with the ground truth regardless the rotation and scaling, which proves the model trained on indoor H3.6M dataset can be well generalized to the MPI-INF-3DHP dataset that includes both indoor and outdoor data. Such a generalization comes from the consensus on the 3D concept of human body between indoor and outdoor scenarios.

\section{Conclusion}
2D-3D ambiguity and lack of sufficient well-labeled 3D human poses are two main challenges in lifting the 2D human pose to the 3D pose. This paper focuses on these two problems and proposes to learn the 3D concept of the human body through domain adaptation for alleviating the ambiguity. The proposed 2D pose lifting architecture integrates the supervised training and the semi-supervised training, which enables us to make use of abundant unlabeled 2D poses. Extensive experiments demonstrates the effectiveness of the proposed method in improving the prediction accuracy and model generalization. Actually, the proposed learning architecture also can be applied to other better pose estimation backbone networks to improve their performance. We also hope that our work can bring some inspirations to researchers in solving other regression problems with learning. In the future, we will explore the proposed method on the unsupervised 3D human pose estimation.

\backmatter




\bigskip

\bibliographystyle{sn-basic}
\bibliography{sn-bibliography}

\begin{thebibliography}{41}
\providecommand{\natexlab}[1]{#1}
\providecommand{\url}[1]{{#1}}
\providecommand{\urlprefix}{URL }
\providecommand{\doi}[1]{\url{https://doi.org/#1}}
\providecommand{\eprint}[2][]{\url{#2}}
 \bibcommenthead

\bibitem[{Cao et~al(2017)Cao, Simon, Wei, and Sheikh}]{cao2017realtime}
Cao Z, Simon T, Wei SE, et~al (2017) Realtime multi-person 2d pose estimation
  using part affinity fields. In: Proceedings of the IEEE conference on
  computer vision and pattern recognition, pp 7291--7299

\bibitem[{Chen et~al(2019)Chen, Tyagi, Agrawal, Drover, Stojanov, and
  Rehg}]{chen2019unsupervised}
Chen CH, Tyagi A, Agrawal A, et~al (2019) Unsupervised 3d pose estimation with
  geometric self-supervision. In: Proceedings of the IEEE/CVF Conference on
  Computer Vision and Pattern Recognition, pp 5714--5724

\bibitem[{Csurka(2017)}]{csurka2017domain}
Csurka G (2017) Domain adaptation for visual applications: A comprehensive
  survey. arXiv preprint arXiv:170205374

\bibitem[{Drover et~al(2018)Drover, Chen, Agrawal, Tyagi, and
  Phuoc~Huynh}]{drover2018can}
Drover D, Chen CH, Agrawal A, et~al (2018) Can 3d pose be learned from 2d
  projections alone? In: Proceedings of the European Conference on Computer
  Vision (ECCV) Workshops, pp 0--0

\bibitem[{Fang et~al(2018)Fang, Xu, Wang, Liu, and Zhu}]{fang2018learning}
Fang HS, Xu Y, Wang W, et~al (2018) Learning pose grammar for monocular 3d pose
  estimation. In: Proceedings of the AAAI Conference on Artificial Intelligence

\bibitem[{Gong et~al(2021)Gong, Zhang, and Feng}]{gong2021poseaug}
Gong K, Zhang J, Feng J (2021) Poseaug: A differentiable pose augmentation
  framework for 3d human pose estimation. arXiv preprint arXiv:210502465

\bibitem[{Habibie et~al(2019)Habibie, Xu, Mehta, Pons-Moll, and
  Theobalt}]{habibie2019wild}
Habibie I, Xu W, Mehta D, et~al (2019) In the wild human pose estimation using
  explicit 2d features and intermediate 3d representations. In: Proceedings of
  the IEEE/CVF Conference on Computer Vision and Pattern Recognition, pp
  10,905--10,914

\bibitem[{Hoffman et~al(2018)Hoffman, Tzeng, Park, Zhu, Isola, Saenko, Efros,
  and Darrell}]{hoffman2018cycada}
Hoffman J, Tzeng E, Park T, et~al (2018) Cycada: Cycle-consistent adversarial
  domain adaptation. In: International conference on machine learning, PMLR, pp
  1989--1998

\bibitem[{Ionescu et~al(2014)Ionescu, Papava, Olaru, and
  Sminchisescu}]{h36m_pami}
Ionescu C, Papava D, Olaru V, et~al (2014) Human3.6m: Large scale datasets and
  predictive methods for 3d human sensing in natural environments. IEEE
  Transactions on Pattern Analysis and Machine Intelligence 36(7):1325--1339

\bibitem[{Iqbal et~al(2020)Iqbal, Molchanov, and Kautz}]{iqbal2020weakly}
Iqbal U, Molchanov P, Kautz J (2020) Weakly-supervised 3d human pose learning
  via multi-view images in the wild. In: Proceedings of the IEEE/CVF Conference
  on Computer Vision and Pattern Recognition, pp 5243--5252

\bibitem[{Kanazawa et~al(2018)Kanazawa, Black, Jacobs, and
  Malik}]{kanazawa2018end}
Kanazawa A, Black MJ, Jacobs DW, et~al (2018) End-to-end recovery of human
  shape and pose. In: Proceedings of the IEEE Conference on Computer Vision and
  Pattern Recognition, pp 7122--7131

\bibitem[{Kocabas et~al(2019)Kocabas, Karagoz, and Akbas}]{kocabas2019self}
Kocabas M, Karagoz S, Akbas E (2019) Self-supervised learning of 3d human pose
  using multi-view geometry. In: Proceedings of the IEEE/CVF Conference on
  Computer Vision and Pattern Recognition, pp 1077--1086

\bibitem[{Kostopoulos et~al(2018)Kostopoulos, Karlos, Kotsiantis, and
  Ragos}]{kostopoulos2018semi}
Kostopoulos G, Karlos S, Kotsiantis S, et~al (2018) Semi-supervised regression:
  A recent review. Journal of Intelligent \& Fuzzy Systems 35(2):1483--1500

\bibitem[{Kundu et~al(2020)Kundu, Seth, Jampani, Rakesh, Babu, and
  Chakraborty}]{kundu2020self}
Kundu JN, Seth S, Jampani V, et~al (2020) Self-supervised 3d human pose
  estimation via part guided novel image synthesis. In: Proceedings of the
  IEEE/CVF Conference on Computer Vision and Pattern Recognition, pp 6152--6162

\bibitem[{Li and Lee(2019)}]{li2019generating}
Li C, Lee GH (2019) Generating multiple hypotheses for 3d human pose estimation
  with mixture density network. In: Proceedings of the IEEE/CVF Conference on
  Computer Vision and Pattern Recognition, pp 9887--9895

\bibitem[{Li and Chan(2014)}]{li20143d}
Li S, Chan AB (2014) 3d human pose estimation from monocular images with deep
  convolutional neural network. In: Asian Conference on Computer Vision,
  Springer, pp 332--347

\bibitem[{Li et~al(2020)Li, Ke, Pratama, Tai, Tang, and Cheng}]{li2020cascaded}
Li S, Ke L, Pratama K, et~al (2020) Cascaded deep monocular 3d human pose
  estimation with evolutionary training data. In: Proceedings of the IEEE/CVF
  Conference on Computer Vision and Pattern Recognition, pp 6173--6183

\bibitem[{Li et~al(2019{\natexlab{a}})Li, Yuan, and
  Vasconcelos}]{li2019bidirectional}
Li Y, Yuan L, Vasconcelos N (2019{\natexlab{a}}) Bidirectional learning for
  domain adaptation of semantic segmentation. In: Proceedings of the IEEE/CVF
  Conference on Computer Vision and Pattern Recognition, pp 6936--6945

\bibitem[{Li et~al(2019{\natexlab{b}})Li, Wang, Wang, and
  Jiang}]{li2019boosting}
Li Z, Wang X, Wang F, et~al (2019{\natexlab{b}}) On boosting single-frame 3d
  human pose estimation via monocular videos. In: Proceedings of the IEEE/CVF
  International Conference on Computer Vision, pp 2192--2201

\bibitem[{Luo et~al(2018)Luo, Chu, and Yuille}]{luo2018orinet}
Luo C, Chu X, Yuille A (2018) Orinet: A fully convolutional network for 3d
  human pose estimation. arXiv preprint arXiv:181104989

\bibitem[{Martinez et~al(2017)Martinez, Hossain, Romero, and
  Little}]{martinez2017simple}
Martinez J, Hossain R, Romero J, et~al (2017) A simple yet effective baseline
  for 3d human pose estimation. In: Proceedings of the IEEE International
  Conference on Computer Vision, pp 2640--2649

\bibitem[{Mehta et~al(2017)Mehta, Rhodin, Casas, Fua, Sotnychenko, Xu, and
  Theobalt}]{mehta2017monocular}
Mehta D, Rhodin H, Casas D, et~al (2017) Monocular 3d human pose estimation in
  the wild using improved cnn supervision. In: 2017 international conference on
  3D vision (3DV), IEEE, pp 506--516

\bibitem[{Mehta et~al(2018)Mehta, Sotnychenko, Mueller, Xu, Sridhar, Pons-Moll,
  and Theobalt}]{mehta2018single}
Mehta D, Sotnychenko O, Mueller F, et~al (2018) Single-shot multi-person 3d
  pose estimation from monocular rgb. In: 2018 International Conference on 3D
  Vision (3DV), IEEE, pp 120--130

\bibitem[{Mitra et~al(2020)Mitra, Gundavarapu, Sharma, and
  Jain}]{mitra2020multiview}
Mitra R, Gundavarapu NB, Sharma A, et~al (2020) Multiview-consistent
  semi-supervised learning for 3d human pose estimation. In: Proceedings of the
  IEEE/CVF Conference on Computer Vision and Pattern Recognition, pp 6907--6916

\bibitem[{Moon et~al(2019)Moon, Chang, and Lee}]{moon2019camera}
Moon G, Chang JY, Lee KM (2019) Camera distance-aware top-down approach for 3d
  multi-person pose estimation from a single rgb image. In: Proceedings of the
  IEEE/CVF International Conference on Computer Vision, pp 10,133--10,142

\bibitem[{Newell et~al(2016)Newell, Yang, and Deng}]{newell2016stacked}
Newell A, Yang K, Deng J (2016) Stacked hourglass networks for human pose
  estimation. In: European conference on computer vision, Springer, pp 483--499

\bibitem[{Nie et~al(2020)Nie, Liu, and Liu}]{nie2020unsupervised}
Nie Q, Liu Z, Liu Y (2020) Unsupervised 3d human pose representation with
  viewpoint and pose disentanglement. In: European Conference on Computer
  Vision, Springer, pp 102--118

\bibitem[{Pavlakos et~al(2017)Pavlakos, Zhou, Derpanis, and
  Daniilidis}]{pavlakos2017coarse}
Pavlakos G, Zhou X, Derpanis KG, et~al (2017) Coarse-to-fine volumetric
  prediction for single-image 3d human pose. In: Proceedings of the IEEE
  Conference on Computer Vision and Pattern Recognition, pp 7025--7034

\bibitem[{Pavllo et~al(2019)Pavllo, Feichtenhofer, Grangier, and
  Auli}]{pavllo20193d}
Pavllo D, Feichtenhofer C, Grangier D, et~al (2019) 3d human pose estimation in
  video with temporal convolutions and semi-supervised training. In:
  Proceedings of the IEEE/CVF Conference on Computer Vision and Pattern
  Recognition, pp 7753--7762

\bibitem[{Rhodin et~al(2018)Rhodin, Sp{\"o}rri, Katircioglu, Constantin, Meyer,
  M{\"u}ller, Salzmann, and Fua}]{rhodin2018learning}
Rhodin H, Sp{\"o}rri J, Katircioglu I, et~al (2018) Learning monocular 3d human
  pose estimation from multi-view images. In: Proceedings of the IEEE
  Conference on Computer Vision and Pattern Recognition, pp 8437--8446

\bibitem[{Sharma et~al(2019)Sharma, Varigonda, Bindal, Sharma, and
  Jain}]{sharma2019monocular}
Sharma S, Varigonda PT, Bindal P, et~al (2019) Monocular 3d human pose
  estimation by generation and ordinal ranking. In: Proceedings of the IEEE/CVF
  International Conference on Computer Vision, pp 2325--2334

\bibitem[{Sun et~al(2019)Sun, Xiao, Liu, and Wang}]{sun2019deep}
Sun K, Xiao B, Liu D, et~al (2019) Deep high-resolution representation learning
  for human pose estimation. In: Proceedings of the IEEE/CVF Conference on
  Computer Vision and Pattern Recognition, pp 5693--5703

\bibitem[{Tekin et~al(2016)Tekin, Katircioglu, Salzmann, Lepetit, and
  Fua}]{tekin2016structured}
Tekin B, Katircioglu I, Salzmann M, et~al (2016) Structured prediction of 3d
  human pose with deep neural networks. In: British Machine Vision Conference
  (BMVC), CONF

\bibitem[{Tzeng et~al(2017)Tzeng, Hoffman, Saenko, and
  Darrell}]{tzeng2017adversarial}
Tzeng E, Hoffman J, Saenko K, et~al (2017) Adversarial discriminative domain
  adaptation. In: Proceedings of the IEEE conference on computer vision and
  pattern recognition, pp 7167--7176

\bibitem[{Wandt and Rosenhahn(2019)}]{wandt2019repnet}
Wandt B, Rosenhahn B (2019) Repnet: Weakly supervised training of an
  adversarial reprojection network for 3d human pose estimation. In:
  Proceedings of the IEEE/CVF Conference on Computer Vision and Pattern
  Recognition, pp 7782--7791

\bibitem[{Wang et~al(2019)Wang, Huang, Wang, and Tao}]{wang2019not}
Wang J, Huang S, Wang X, et~al (2019) Not all parts are created equal: 3d pose
  estimation by modeling bi-directional dependencies of body parts. In:
  Proceedings of the IEEE/CVF International Conference on Computer Vision, pp
  7771--7780

\bibitem[{Wei et~al(2016)Wei, Ramakrishna, Kanade, and
  Sheikh}]{wei2016convolutional}
Wei SE, Ramakrishna V, Kanade T, et~al (2016) Convolutional pose machines. In:
  Proceedings of the IEEE conference on Computer Vision and Pattern
  Recognition, pp 4724--4732

\bibitem[{Yang et~al(2018)Yang, Ouyang, Wang, Ren, Li, and Wang}]{yang20183d}
Yang W, Ouyang W, Wang X, et~al (2018) 3d human pose estimation in the wild by
  adversarial learning. In: Proceedings of the IEEE Conference on Computer
  Vision and Pattern Recognition, pp 5255--5264

\bibitem[{Zhao et~al(2019)Zhao, Peng, Tian, Kapadia, and
  Metaxas}]{zhao2019semantic}
Zhao L, Peng X, Tian Y, et~al (2019) Semantic graph convolutional networks for
  3d human pose regression. In: Proceedings of the IEEE/CVF Conference on
  Computer Vision and Pattern Recognition, pp 3425--3435

\bibitem[{Zhou et~al(2019)Zhou, Han, Jiang, Jia, and Lu}]{zhou2019hemlets}
Zhou K, Han X, Jiang N, et~al (2019) Hemlets pose: Learning part-centric
  heatmap triplets for accurate 3d human pose estimation. In: Proceedings of
  the IEEE/CVF International Conference on Computer Vision, pp 2344--2353

\bibitem[{Zhou et~al(2017)Zhou, Huang, Sun, Xue, and Wei}]{zhou2017towards}
Zhou X, Huang Q, Sun X, et~al (2017) Towards 3d human pose estimation in the
  wild: a weakly-supervised approach. In: Proceedings of the IEEE International
  Conference on Computer Vision, pp 398--407

\end{thebibliography}




\end{document}